\pdfoutput=1

\documentclass[11pt]{article}

\usepackage{acl}

\usepackage{times}
\usepackage{latexsym}

\usepackage[T1]{fontenc}

\usepackage[utf8]{inputenc}

\usepackage{microtype}

\usepackage{amsmath,amssymb}
\usepackage{enumitem}
\usepackage{booktabs}
\usepackage{multirow}
\usepackage{graphicx}
\usepackage{svg}

\definecolor{darkpastelgreen}{rgb}{0.01, 0.75, 0.24}
\definecolor{darkenedred}{rgb}{0.8, 0.0, 0.0}
\newcommand{\better}[1]{{\textcolor{darkpastelgreen}{#1}}}
\newcommand{\worse}[1]{{\textcolor{darkenedred}{#1}}}

\title{Predicting Fine-Tuning Performance with Probing}

\author{Zining Zhu$^{1,2}$, Soroosh Shahtalebi$^{2}$, Frank Rudzicz$^{1,2,3}$\\
  $^1$ University of Toronto  $^2$ Vector Institute for Artificial Intelligence $^3$ Unity Health Toronto \\
  \texttt{zining@cs.toronto.edu, soroosh.shahtalebi@vectorinstitute.ai} \\ \texttt{frank@spoclab.com} \\}

\begin{document}
    \maketitle
\begin{abstract}
Large NLP models have recently shown impressive performance in language understanding tasks, typically evaluated by their fine-tuned performance. Alternatively, probing has received increasing attention as being a lightweight method for interpreting the intrinsic mechanisms of large NLP models. In probing, post-hoc classifiers are trained on ``out-of-domain'' datasets that diagnose specific abilities. While probing the language models has led to insightful findings, they appear disjointed from the development of models. This paper explores the utility of probing deep NLP models to extract a proxy signal widely used in model development -- the fine-tuning performance. We find that it is possible to use the accuracies of only three probing tests to predict the fine-tuning performance with errors $40\%$ - $80\%$ smaller than baselines. We further discuss possible avenues where probing can empower the development of deep NLP models.
\end{abstract}

\section{Introduction}
Large-scale neural models have recently demonstrated state-of-the-art performance in a wide variety of tasks, including sentiment detection, paraphrase detection, linguistic acceptability, and entailment detection \citep{devlin-etal-2019-bert,radford2019language,peters-etal-2018-deep}. Developing systems for these tasks usually involves two stages: a pre-training stage, where the large neural models gain linguistic knowledge from weak supervision signals in massive corpora, and a fine-tuning stage, where the models acquire task-specific knowledge from labeled data. The fine-tuning results are widely used to benchmark the performances of neural models and refine the models' development procedures.

However, these fine-tuning results are summary statistics and do not paint the full picture of deep neural models \citep{ethayarajh-jurafsky-2020-utility,bender-koller-2020-climbing}. As researchers are increasingly concerned about interpreting the intrinsic mechanisms of deep neural models, many data-driven assessment methods have been developed. These assessments usually follow the route of compiling a targeted dataset and running post-hoc analyses. Until now, one of the most popular interpretation methods is referred to as {\em probing}. To probe a neural model, one uses a predictor to obtain the labels from the representations that are embedded using the neural model. Probing analyses on deep neural models revealed some low-dimensional syntactic structures \citep{hewitt-manning-2019-structural}, common-sense knowledge \citep{petroni-etal-2019-language} and (to some extent) human-like abilities, including being surprised upon witnessing linguistic irregularity \citep{li-etal-2021-bert} and reasoning about space and time \citep{aroca-ouellette-etal-2021-prost}.

From the viewpoint of data-driven assessments, both fine-tuning and probing can reveal the abilities of deep neural networks, but they appear to steer towards different directions:

\textit{In-domain vs. out-of-domain.} Fine-tuning uses in-domain data -- we evaluate the models on the same distributions as those in deployment. Probing, however, uses out-domain data: instead of simulating the deployment environment, the targeted datasets focus on diagnosing specific abilities. 

\textit{Inclusive vs. specific.} In fine-tuning, edge cases should be included, so the unexpected behavior after deployment can be minimized \citep{ribeiro-etal-2020-beyond} and the fine-tuning results can be stable \citep{zhang2021revisiting}. On the contrary, the probing datasets are more specialized, so smaller datasets suffice.\footnote{Another viewpoint for the dataset requirement can be derived from learning theory. Loosely speaking, optimizing more parameters requires more data to reach stable results. Fine-tuning involves more parameters than probing. \citet{probing_dataset} provides a more quantitative discussion.}

\textit{High performances vs. faithful interpretations.} While fine-tuning methods are mainly studied from an algorithmic perspective to enhance the performance of language models, probing methods aim at assessing the faithfulness of language models. To fulfill the former objective, fine-tuning is accompanied by efforts in pre-training, collecting more data, building better representations, and exploring novel model architectures \citep{he_deberta_2021,sun2021ernie,wang2021towards,jiang-etal-2020-smart}. Conversely, the latter goal is pursued by borrowing inspirations from a variety of other sources, including psycholinguistic assessment protocols \citep{futrell-etal-2019-neural,li-etal-2022-neural}, information theory \citep{voita_information-theoretic_2020,pimentel-cotterell-2021-bayesian,zhu-rudzicz-2020-information}, and causal analysis \citep{slobodkin-etal-2021-mediators,elazar_amnesic_2021}.

In short, probing assessments are more specialized (therefore more flexible) and less computationally expensive. In contrast, the performance scores of fine-tuning assessments are more relevant to the design and training of deep neural models. \textbf{Can probing be used in the development of deep neural models?} This question involves two aspects: 
\begin{itemize}[nosep]
    \item \textit{Feasibility}: Are probing results relevant in the model development?
    \item \textit{Operation}: How to set up probing analyses to get these useful results?
\end{itemize}

This paper attempts to answer both. For feasibility, we show that a crucial feedback signal in model development, the fine-tuning performance, can be predicted via probing results, indicating a positive answer to the feasibility question.

For operation, we run extensive ablation studies to simplify the probing configurations, leading to some heuristics to set up probing analyses. We start with a battery of probing tasks and evaluate the utilities both task-wise and layer-wise (\S \ref{subsec:exp:one_probing_task} - \S \ref{subsec:exp:anova_layers}). We then reduce the number of probing configurations, showing that as few as 3 configurations can predict fine-tuning results with RMSEs between $40\%$ and $80\%$ smaller than the control baseline (\S \ref{subsec:exp:best_3_features}). To further answer the operation question, we run ablation studies on different probing configurations, including probing methods (\S \ref{subsec:exp:ablation_probing_configuration}) and the number of data samples (\S \ref{subsec:exp:ablation_smaller_probing_datasets}). We also analyze the uncertainty of the results (\S \ref{subsec:exp:ablation-random-seed}). Our analysis shows the possibility of using probing in developing high-performance deep neural models.

All codes are open-sourced at \url{https://github.com/SPOClab-ca/performance_prediction}.

\section{Related Work} 

\paragraph{Performance prediction}
\citet{xia-etal-2020-predicting} proposed a framework that predicts task performance using a collection of features, including the hyperparameters of the model and the percentage of text overlap between the source and target datasets. \citet{srinivasan2021predicting} extended this framework into a multilingual setting. \citet{ye-etal-2021-towards} considered the reliability of performance -- an idea similar to that of \citet{dodge-etal-2019-show}. This paper differs from the performance prediction literature in the set of features -- we use the probing results as features -- and more importantly, we aim at showing that the probing results can improve the interpretability in the development procedures of large models.

\paragraph{Out-of Domain generalization} The out-of-domain generalization literature provides a variety of methods to improve the performance of out-of-domain classification. We defer to \citet{wang2021generalizing} for a summary. \citet{gulrajani2020search} ran empirical comparisons on many algorithms, and some theoretical analyses bound the performance of out-of-domain classification \citep{li2022finding,minsker2019excess}. In our setting, the probing and the fine-tuning datasets can be considered different domains, but our analysis predicts the out-of-domain performance. A similar setting was presented in \citet{kornblith_better_2019}, which studied the correlation between the performance on ImageNet and the performance of transfer learning on a variety of image domains. Our setting focuses on text domains, and use specialized, small-sized probing datasets.

\paragraph{Probing, and the utility of LODNA} The probing literature reveals various abilities of deep neural models, as summarized by \citet{rogers_primer_2020,manning_emergent_2020,belinkov_probing_2021,pavlick_semantic_2022}. 
There have been some discussions on the utility of probing results. \citet{baroni_proper_2021} argued that these linguistic-oriented deep neural network analyses (LODNA) should treat deep neural models as algorithmic linguistic theories; otherwise, LODNA has limited relevance to theoretical linguists. 
Recent literature in LODNA drew interesting findings by comparing the mechanisms in which algorithms and humans respond to external stimuli, including the relative importance of sentences \citep{hollenstein-beinborn-2021-relative}. 
Probing results, when used jointly with evidence from datasets, can also be used to predict the inductive bias of neural models \citep{lovering_predicting_2021,immer2021probing}.
As we show, probing results can explain the variance in and even predict the fine-tuning performance of neural NLP models.

\paragraph{Fine-tuning and probing}
There have been multiple papers that explored fine-tuning and probing paradigms. Probing is used as a post-hoc method to interpret linguistic knowledge in deep neural models during pre-training \citep{liu_linguistic_2019}, fine-tuning \citep{miaschi_linguistic_2020,mosbach-etal-2020-interplay-fine,durrani-etal-2021-transfer,yu-ettinger-2021-interplay,zhou2021closer}, and other stages of model development \citep{Ebrahimi2021HowDA}. From a performance perspective, probing can sometimes result in higher performance metrics (e.g., accuracy) than fine-tuning \citep{liu_linguistic_2019,hall_maudslay_tale_2020} and fine-tuning can benefit from additional data \citep{phang2018sentence}. We take a different perspective, considering how the probing and fine-tuning results relate to each other, and more importantly, how the signals of probing can be helpful towards developing large neural models.

\section{Methods}
We present the overall analysis method and evaluation metric in this section. \S \ref{sec:experiments} elaborates the detailed experiment settings.

\paragraph{Predicting fine-tuning performance}
A deep neural model $M$ can be fine-tuned on task $t$ to achieve performance $\mathcal{A}_t$. Let $\mathbf{S} \in \mathbb{R}^{N}$ be the test accuracies of probing classifications on model $M$, using $N$ configurations. 
For example, a deep neural model $M=\textrm{RoBERTa}$ can be fine-tuned to reach performance $\mathcal{A}_t=0.85$ on a $t=\textrm{RTE}$ task. With post-hoc classifiers applied to the 12 layers of $M$, we can probe for 12 test accuracies on a probing task (e.g., detecting the past vs. present tense), which constitute of $\mathbf{S}$.\footnote{Following the default implementation of linear regression, we include an additional dimension in $\mathbf{S}^{(k)}$ to multiply with the bias term, so $\mathbf{S}^{(k)}\in \mathbb{R}^{N+1}$ in the following equations.}

To find the pattern across a diverse category of models, we regress over $K$ models (we will describe in \S \ref{subsec:pretrained-models}). The collected probing results $\{\mathbf{S}^{(k)}\}_{k=1}^{K}$ can be used to predict the fine-tuning performance $\{\mathcal{A}_t^{(k)} \}_{k=1}^{K}$ via regression. Formally, this procedure optimizes for $N+1$ parameters, $\mathbf{\theta} \in \mathbb{R}^{N+1}$ so that:
\begin{equation}
    \mathbf{\theta_*} = \text{argmin}_{\mathbf{\theta}} \Sigma_k ||\mathbf{\theta}^T\mathbf{S}^{(k)} - \mathcal{A}_t^{(k)}||^2
\end{equation}

This procedure has closed-form solutions that are implemented in various scientific computation toolkits (e.g., \texttt{R} and \texttt{scipy}). The minimum reachable RMSE is therefore:
\begin{equation}
    \text{RMSE} = \sqrt{\frac{1}{K}\Sigma_k ||\mathbf{\theta_*}^T\mathbf{S}^{(k)} - \mathcal{A}_t^{(k)}||^2}
\end{equation}

\paragraph{RMSE-reduction}
While RMSE can evaluate the quality of this regression, it is insufficient for measuring the informativeness of $\mathbf{S}$ due to the discrepancy among the fine-tuning tasks $t$. Suppose we have two tasks, $t_1$ and $t_2$, where the probing results $\mathbf{S}$ can support high-precision regressions to $\textrm{RMSE}=0.01$ on both tasks. However, on $t_1$, even features drawn from random distributions\footnote{Considering the small data sizes (i.e., the total number of models studied), even the ``random features'' drawn from random noises contain artefacts -- patterns that can be used to regress the results.} might be sufficient to reach $\textrm{RMSE}=0.02$, while on the more difficult task, $t_2$, random features could only reach $\textrm{RMSE}=0.10$ maximum. The probing results $\mathbf{S}$ is more useful for $t_2$ than $t_1$, but RMSE itself does not capture this difference.

Considering this, we should further adjust against a baseline, the minimum reachable RMSE using random features.
\begin{equation}
    \theta_{c*} = \text{argmin}_{\mathbf{\theta}} \Sigma_k ||\mathbf{\theta}^T\mathbf{\epsilon}^{(k)} - \mathcal{A}_t^{(k)}||^2,
\end{equation}
where the random features $\mathbf{\epsilon}$ are drawn from  $\mathcal{N}(0, 0.1)$. Overall, the RMSE and the reduction from the baseline are computed as:
\begin{align}
    \textrm{RMSE}_c = \sqrt{\frac{1}{K}\Sigma_k ||\mathbf{\theta}_{c*}^T\mathbf{\epsilon}^{(k)} - \mathcal{A}_t^{(k)}||^2} \\
    \textrm{RMSE\_reduction} =\frac{\textrm{RMSE}_c - \textrm{RMSE}}{\textrm{RMSE}_c} \times 100
\end{align}
In the experiments, all RMSE and $\text{RMSE}_c$ values follow 5-fold cross validation. We report the RMSE\_reduction as the score that measures the utility of $\mathbf{S}$.

\section{Evaluation tasks and datasets}
\subsection{Fine-tuning tasks}
\label{subsec:fine-tuning-tasks}
We consider 6 binary classification tasks in GLUE \citep{wang_glue_2019} as fine-tuning tasks:
\texttt{RTE} consists of a collection of challenges recognizing textual entailment. Given two sentences, the model decides whether a sentence entails the other.
\texttt{COLA} \citep{warstadt2019neural} requires the model to determine if a sentence is linguistically acceptable.
\texttt{MRPC} \citep{dolan-brockett-2005-automatically} requires the model to identify if a pair of sentences are paraphrases. 
\texttt{SST2} \citep{socher-etal-2013-recursive} asks the model to output the sentiment positivity of movie reviews.
\texttt{QNLI} contains questions and answers parsed from SQuAD \citep{rajpurkar-etal-2016-squad}. This task requires the model to decide whether the \textit{answer} answers the \textit{question}.
\texttt{QQP}\footnote{\url{https://quoradata.quora.com/First-Quora-Dataset-Release-Question-Pairs}} tests if the model can correctly output whether a pair of Quora questions are synonymous.

\subsection{Probing tasks}
\label{subsec:probing-tasks}
We use 7 probing tasks from SentEval \citep{conneau_senteval_2018} which can be approximately grouped in two categories, syntactic and semantic:
\begin{itemize}[nosep]
    \item Syntactic: bigram shift (BShift), and tree depth (TreeDepth) 
    \item Semantic: past present (Tense), subject number (SubjNum), object number (ObjNum), semantic odd-man out (SOMO), and coordination inversion (CoordInv)
\end{itemize}
These probing tasks span across a range of linguistic abilities. In general, layers closer to the inputs (lower layers) in BERT contain more surface-level information, whereas higher layers contain more syntactic and semantic information \citep{tenney_bert_2019,jawahar_what_2019}, but the actual location of different linguistic features may vary \citep{miaschi_linguistic_2020}.
The SentEval datasets are usually hundreds of times larger than what would be sufficient to support statistically significant comparisons \citep{probing_dataset}, so we randomly sample 1200 data points per class, corresponding to around 1\% of the original SentEval data.

\subsection{Pre-trained Language Models}
\label{subsec:pretrained-models}
We use several most widely used pre-trained language models for fine-tuning and probing. We refer to the models by their names on the Huggingface Model Hub.\footnote{\url{https://huggingface.co/}}

\texttt{roberta-base} \citep{liu_roberta_2019} pretrains BERT \citep{devlin-etal-2019-bert} on over 160GB of English corpora, using improved techniques including dynamic masking, large mini-batches and masked language modeling without next-sentence-prediction. 

\texttt{xlm-roberta-base} \citep{conneau-etal-2020-unsupervised} is pre-trained on 2.5TB of Common Crawl data from over 100 languages. The multiple languages sources improve the transferability across languages while compromising only a little accuracy on the English GLUE tasks (compared to the monolingual RoBERTa).

\texttt{albert-base-v2} \citep{Lan2020ALBERT} shares parameters across layers and decomposes the vocabulary matrices into smaller matrices. These parameter-reducing techniques reduce the computation resource requirements, which allows the model pretraining to further scale-up.

\texttt{microsoft/deberta-base} \citep{he_deberta_2021} uses separate attention vectors to model the content and the positions of each word. During fine-tuning, DeBERTa adds adversarial perturbations to the normalized embeddings.

\texttt{xlnet-base-cased} \citep{yang_xlnet_2019} models different permutation orders of the contexts during pre-training. XLNet additionally uses attentions to keep track of previous states, allowing the model to process the contexts extending beyond fixed lengths.

\texttt{Corrupted models}. To increase the diversity of models, we corrupt the language models on a masked language modeling task by MLM fine-tuning on scrambled Wikipedia\footnote{\texttt{wikitext-2-v1} from huggingface \texttt{datasets}.} for 500, $1k$, $2k$, $4k$, and $6k$ steps. This ``model augmentation'' procedure does not apply to XLNet because scrambling the corpus produces a permutation of context, which XLNet already models. In total, there are 25 language models, each containing 12 layers.

\subsection{Fine-tuning methods}
\label{subsec:fine-tuning-methods}
For all fine-tuning classifications, we use the \textit{AutoModelForSequenceClassification} framework by huggingface Transformers \citep{wolf-etal-2020-transformers}. The model is trained with an AdamW optimizer with a collection of initial learning rates\footnote{1e-4, 3e-5, 1e-5, 3e-6, 1e-6, 3e-7. Note that most highest-performing classification runs are reached with either 1e-5 or 3e-6.} and a batch size of 4. Since the GLUE tasks do not publicize the test set labels, we use the best dev set performance as the fine-tuning results. For reproducibility, we fix the random seed to 42 in PyTorch \citep{paszke_pytorch_2019}. Additional details, including runtime and computation resources, are in Appendix \ref{subsec:computation}.

\subsection{Probing methods}
\label{subsec:probing-methods}
There are many methods to probe a neural network. In this paper, we use a post-hoc classifier to predict a target (``probing task target'') from the representations of the first token (CLS). We run through a collection of scikit-learn \citep{pedregosa_scikit-learn_nodate} classifiers,\footnote{Logistic Regression, MLP with 10 and 20 hidden nodes, Random Forest with 100 and 10 estimators, Decision Tree, and SVM. There are 7 classifiers in total.} choose the best one by the dev accuracy, and take its test accuracy as the probing result $\mathbf{S}$. Additional details, including runtime and computation resources, are in Appendix \ref{subsec:computation}.

\section{Experiments}
\label{sec:experiments}
\subsection{Fine-tuning performance}
\label{subsec:exp:finetuning_performances}
\begin{figure}[t]
    \centering
    \includesvg[width=\linewidth]{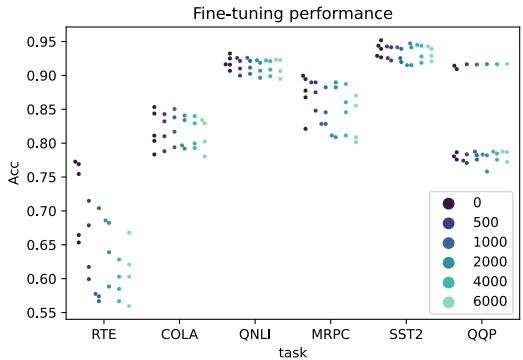}
    \caption{Fine-tuning performance. The color coding reflects the number of corruption steps on scrambled wikipedia, with 0 corresponding to the ``vanilla'' language models.}
    \label{fig:glue_accuracies}
\end{figure}

As an exploratory analysis, Figure \ref{fig:glue_accuracies} plots the distributions of the GLUE fine-tuning performances. The additional corruption steps result in more significant fine-tuning performance drops on RTE and MRPC than other tasks. Moreover, on QQP, the dev accuracies of \texttt{roberta-base} (and its corrupted models) are larger than 0.90, while most other models have around 0.80 dev accuracies.

\subsection{Which probing task is most informative?}
\label{subsec:exp:one_probing_task}
We start with testing the predictability of using the results from only one probing task. For each probing task, we concatenate the 12 probing results as features and predict the fine-tuning performance using linear regression.\footnote{\texttt{lm} method in the \texttt{caret} R package.}

Table \ref{tab:rmse-results} shows the percentage of RMSE reduction from baseline, using all layers from one probing task. There is no definitive answer towards ``which probing task best predicts all fine-tuning tasks'' but, depending on the linguistic abilities that each task targets, there are some regularities.
For example, the `number counting' probing tasks do not predict the fine-tuning performances on RTE, the textual entailment recognition task. In other fine-tuning tasks (COLA, QNLI, MRPC, SST2, QQP), however, each probing task shows positive RMSE reduction, signaling the ability to predict fine-tuning performance.

\begin{table*}[t]
    \centering
    \resizebox{\textwidth}{!}{
    \begin{tabular}{r| r r r r r r | r}
        \toprule 
         & RTE & COLA & MRPC & SST2 & QNLI  & QQP & \textit{Average} \\ \midrule 
        All layers one task (\S \ref{subsec:exp:one_probing_task}) &&&&&& \\
        BShift & 6.24 & 52.80 & \textbf{53.18} & 29.78 & 55.29 & 51.64 & \hspace{1em}41.49 \\
        CoordInv & 2.10 & 66.59 & 18.18 & 44.24 & 56.35 & 56.57 & 40.67 \\
        ObjNum & 2.19 & 44.20 & 28.02 & 53.15 & 60.64 & 72.38 & 43.33 \\
        SOMO & 30.90 & 44.75 & 29.39 & 29.28 & 38.64 & 55.68 & 38.11 \\
        Tense & 3.07 & 48.42 & 34.65 & 22.29 & 41.37 & 75.58 & 37.56 \\
        SubjNum & -19.66 & \textbf{78.56} & 34.48 & 47.75 & 64.74 & 51.50 & 42.90 \\
        TreeDepth & 4.37 & 53.03 & 9.54 & 46.98 & 62.79 & 54.67 & 38.56 \\ \midrule 
        One layer per task (\S \ref{subsec:exp:one_layer_per_task}) & 36.12 & 62.66 & 25.78 & 49.87 & 59.79 & 26.73 & 43.49 \\ \midrule
        \multirow{2}{*}{Only three features (\S \ref{subsec:exp:best_3_features})} & \textbf{41.69} & 75.66 & 47.56 & \textbf{72.59} & \textbf{80.52} & \textbf{76.77} & \textbf{65.80} \\
        & CoordInv\_1 & ObjNum\_2 & TreeDepth\_1 & SubjNum\_1 & SubjNum\_2 & TreeDepth\_6 & N/A \\
        & TreeDepth\_1 & SubjNum\_2 & SOMO\_4 & BShift\_ 3 & Tense\_8 & Tense\_8 & N/A \\
        & BShift\_12 & TreeDepth\_12 & ObjNum\_7 & CoordInv\_10 & CoordInv\_9 & Tense\_12 & N/A \\
        \bottomrule
    \end{tabular}}
    \caption{RMSE reduction from baseline. A larger value shows the probing results more indicative of the fine-tuning performance. A small (or even negative) value means the probing results are not informative, compared to random features. The \textbf{bold-font} configurations are those with the highest RMSE reductions for predicting each fine-tuning task (i.e., within each column).}
    \label{tab:rmse-results}
\end{table*}

\subsection{Which layers are the most indicative?}
\label{subsec:exp:anova_layers}
In the regression experiments of \S \ref{subsec:exp:one_probing_task}, we considered each feature equally important. However, a one-way ANOVA shows that some layers are more indicative than others, as Table \ref{tab:significant-reatures-one-probing-task} shows. For example, the probing results of tree\_depth (at layer 1) and object\_number (at layer 1) explain significant variance on all fine-tuning tasks.

Note that the layers with the most predictability should not be confused with those containing the richest linguistic knowledge. The former corresponds to the probing results that explain the most variances, while the latter corresponds with probing with the highest accuracy. 


\begin{table*}[t]
    \centering
    \resizebox{0.9\textwidth}{!}{
    \begin{tabular}{r| r r r r r r r}
        \toprule 
         & RTE & COLA & MRPC & SST2 & QNLI & QQP \\ \midrule 
        bigram shift (BShift) & 4,5 & 2,4,5 & 2,4,5,9 & 2,5,6 & 2,4,5 & 2,4,5 \\
        coordination inversion (CoordInv) & 5,6,12 & 1,2,4,6 & 1,6 & 1,4,6 & 1,4-6 & 2-4,6 \\
        object number (ObjNumber) & 1 & 1,3,8,11 & 1,3 & 1,3-5,8,11 & 1,3,8,11 & 1-5,12 \\
        semantic odd man out (SOMO) & 4,5,8,12 & 2-6 & 3,4 & 3,5,6 & 2-6 & 2,5-9,12 \\
        past present (Tense) & 1 & 1,3,5 & 1,5,6 & 1,11 & 1,3,5,8 & 1-5,8-11 \\
        subject number (SubjNum) & None & 1,3-6,9 & 1 & 1,4 & 1 & 1,2,3,4 \\
        tree depth (TreeDepth) & 1 & 1 & 1 & 1,3,5 & 1 & 1-3,7,8,11 \\
        \bottomrule
    \end{tabular}}
    \caption{Layers with significant probing results ($p<.05$ from one-way ANOVA) with residual $\textrm{dof}=12$.}
    \label{tab:significant-reatures-one-probing-task}
\end{table*}

\subsection{Only one layer per probing task}
\label{subsec:exp:one_layer_per_task}
Instead of probing all 12 layers, could using the probing results from only one layer for each probing task be beneficial? Following Table \ref{tab:significant-reatures-one-probing-task}, we use the layers that are shown to explain significant portions of variance for the most fine-tuning tasks.\footnote{Namely, layers 5, 6, 1, 5, 1, 1, 1 from the 7 probing tasks respectively.}

The results are also included in Table \ref{tab:rmse-results}. When reducing the number of features into around half (12 to 7), ``one layer per probing task'' can reduce more RMSE in RTE and SST2.\footnote{Note that the ANOVA in \S \ref{subsec:exp:anova_layers} use all data samples, so the choice of the features contain information propagated from the validation set. To ensure fair comparisons, we do \textit{not} include the results from ``one-layer-per-task'' setting when finding the highest RMSE reductions in subsequent analysis. The results from this setting do not outperform the bold-font results even once, though.} However, the results in other fine-tuning tasks indicate that alternative feature selection methods might help find a more predictive feature set.

\subsection{Can we predict with only 3 features?}
\label{subsec:exp:best_3_features}
This experiment further reduces the number of features used while maximizing the MSE reductions. We iterate through all possible combinations of the $12\times 7=84$ probing features for each fine-tuning task and report the largest RMSE reduction in predicting the fine-tuning performance.

Table \ref{tab:rmse-results} shows the results and the corresponding features. The prediction with three features can reduce the most RMSE on RTE, SST2, QNLI, and QQP. On COLA and MRPC, the RMSE reductions by the top three features are at most 6\% smaller than those of the best previous configurations, which involved many more probing features.

The results show the utility of probing. It is possible to predict the fine-tuning performances by probing on as few as three configurations (each configuration using one probing task on one layer).

\begin{table*}
\centering 
\resizebox{.9\textwidth}{!}{
\begin{tabular}{r | r r r r r r | r}
    \toprule 
     & RTE & COLA & MRPC & SST2 & QNLI & QQP & \textit{Average} \\ \midrule 
    Highest-accuracy probe in \S \ref{subsec:exp:one_probing_task} - \S \ref{subsec:exp:best_3_features} & 41.69 & 78.56 & 53.18 & \textbf{72.59} & 80.52 & 76.77 & 67.22 \\ \midrule 
    Specify one probing method (\S \ref{subsec:exp:ablation_probing_configuration}) &&&&&& \\
    DecisionTree & 51.98 & 68.48 & 54.31 & 70.90 & 74.35 & 52.85 & 62.15 \\
    LogReg & 45.28 & 78.34 & 44.87 & 70.26 & 83.13 & 73.98 & 65.98 \\
    MLP-10 & 48.50 & 72.12 & 45.88 & 65.87 & 73.82 & 81.97 & 65.69 \\
    MLP-20 & 47.37 & 74.94 & \textbf{63.79} & 69.22 & 79.10 & \textbf{82.67} & \textbf{69.52} \\
    RandomForest-10 & 50.64 & 74.08 & 50.17 & 68.20 & 75.19 & 59.66 & 62.99 \\
    RandomForest-100 & \textbf{53.94} & \textbf{79.20} & 53.21 & 71.60 & \textbf{83.25} & 72.72 & 68.99 \\
    SVM & 51.71 & 74.01 & 57.92 & 71.44 & 76.78 & 73.03 & 67.48 \\
    \bottomrule 
\end{tabular}}
\caption{Maximum RMSE reductions using different probing configurations. The \textbf{bold-font} numbers are the maximum values in each column.}
\label{tab:comparing_probing_configurations}
\end{table*}

\subsection{Ablation: probing configuration}
\label{subsec:exp:ablation_probing_configuration}
To further simplify the probing procedure, we run this ablation study. Instead of probing using a battery of post-hoc classifiers (as mentioned in \S \ref{subsec:probing-methods}), we test if the probing results from each individual classifier can reproduce the findings of \S \ref{subsec:exp:one_probing_task} - \S \ref{subsec:exp:best_3_features}.

Table \ref{tab:comparing_probing_configurations} shows the maximum MSE reductions using different choices of probes. A perhaps surprising finding is that the probes selected from the ``highest-accuracy'' criterion do not always produce the most valuable results. To predict fine-tuning performances, directly specifying the probing method as MLP-20 or RandomForest-100 may be instead more recommended.

As a side note, among the 48 results presented in Table \ref{tab:comparing_probing_configurations}, only 9 are not achieved by the ``best-3-features'' methods (including the 2 shown in Table \ref{tab:rmse-results}). This contrast emphasizes the importance of feature selection when configuring probes.

\subsection{Ablation: dataset size}
\label{subsec:exp:ablation_smaller_probing_datasets}
The findings in \S \ref{subsec:exp:one_probing_task} - \S \ref{subsec:exp:ablation_probing_configuration} show that as few as 1,200 samples per class (around 1\% of total data) are sufficient to provide useful findings. What if we further reduce the sizes of probing datasets? Here, we repeat \S \ref{subsec:exp:one_probing_task} and \S \ref{subsec:exp:best_3_features} with probing results from only 400 samples per class. While we can also reduce RMSE with only 400 samples per class, probing results are generally not as useful as those from 1,200 samples. Among the 48 configurations, the probing results from 400 samples have worse RMSE reductions in 11 configurations, but better in 5. Detailed results are included in Table \ref{tab:rmse-results-400perclass}. 

\subsection{Uncertainty analysis}
\label{subsec:exp:ablation-random-seed}
Our method involves comparing the maximum RMSE reductions against the baseline (regressing from features drawn from Gaussian) $\text{RMSE}_c$, which may be affected by the random seeds. Here we describe an error analysis on the baseline regressor results of \S \ref{subsec:exp:one_probing_task} - \S \ref{subsec:exp:best_3_features}.

We run $N=100$ Monte Carlo simulations on each configuration of regression from 3, 7, and 12 features, respectively, record the $\textrm{RMSE}_{c}$, and analyze the uncertainty. We use the variation of $\textrm{RMSE}_c$ (as measured by $\text{Std}(\textrm{RMSE}_c)$) relative to the scale (as measured by $\textrm{Mean}(\textrm{RMSE}_c)$) to describe the uncertainty. As shown in Table  \ref{tab:std_to_mean_ratio_mse_c}, the uncertainty remains relatively stable across the choice of regression tasks but increases with the number of features. This result favors the use of fewer probing results as features.

Note that these uncertainty values are nontrivial. Let us take COLA as an example. To regress the fine-tuning performance, a 3-feature setting can achieve 75.66\% RMSE reduction compared to $\text{RMSE}_c$, but $\text{RMSE}_c$ itself has 5.46\% uncertainty. This translates to around 7.22\% uncertainty for the RMSE-reduction results (Table \ref{tab:rmse-results}). 

Can we reduce the uncertainty by using alternative evaluation metrics like the RMSE, or the percentage of explained variance (\texttt{ExplVar}), instead of introducing a control task? In addition to the adjustment for dataset artifacts, the control task provides a baseline to understand the utility. While the RMSE is always positive and \texttt{ExplVar} is almost always above 90\%, RMSE reduction itself provides a clearer picture of the utility of probing features.

\begin{table*}[t]
\centering
    \resizebox{.9\textwidth}{!}{
    \begin{tabular}{r| r r r r r r | r}
        \toprule 
         & RTE & COLA & MRPC & SST2 & QNLI  & QQP & \hspace{.5em}\textit{Average} \\ \midrule
        All layers one task &&&&&& \\
        BShift & 21.00 & 53.66 & \worse{19.65} & 35.60 & 51.32 & 57.55 & 39.80 \\
        CoordInv & 4.49 & \worse{31.28} & 22.83 & 38.93 & \worse{6.30} & \worse{25.51} & \worse{21.56} \\
        ObjNum & \better{29.51} & 56.10 & 39.94 & 65.95 & 72.30 & 70.54 & 55.72 \\
        SOMO & \worse{-5.09} & \worse{-7.14} & 16.09 & \worse{9.36} & \worse{-0.71} & 46.36 & \worse{9.81} \\
        Tense & 1.30 & 51.79 & 33.63 & 13.52 & 49.43 & 73.01 & 37.11 \\
        SubjNum & \better{9.89} & 76.32 & 47.19 & 48.62 & 65.36 & 49.42 & 49.47\\
        TreeDepth & \worse{-11.49} & 66.06 & \better{28.98} & \worse{27.13} & 59.93 & 43.10 & 35.62 \\ \midrule 
        \multirow{2}{*}{Only three features} & \better{47.38} & 77.84 & \better{56.70} & 72.27 & 82.01 & \worse{71.08} & 67.88 \\
         & Tense\_1 & SubjNum\_1 & BShift\_2 & SubjNum\_1 & SubjNum\_1 & Tense\_3 & N/A\\
         & SubjNum\_11 & BShift\_6 & ObjNum\_7 & Tense\_2 & SubjNum\_8 & BShift\_4 & N/A\\
         & CoordInv\_12 & TreeDepth\_8 & SOMO\_9 & CoordInv\_6 & BShift\_9 & BShift\_8 & N/A\\
        \bottomrule
    \end{tabular}}
    \caption{RMSE reduction from baseline, using probing results with 400 data samples per class. The colored results are different from (\better{better} than or \worse{worse} than) the results with 1,200 data samples (Table \ref{tab:rmse-results}) by more than the estimated uncertainty margins in \S \ref{subsec:exp:ablation-random-seed}, i.e., 5\% and 15\% for 3 and 12 features, respectively.}
    \label{tab:rmse-results-400perclass}
\end{table*}

\begin{table}[t]
    \centering
    \begin{tabular}{r| r r r}
        \toprule 
         & 3 features & 7 features & 12 features \\ \midrule 
        RTE & 5.71\% & 9.00\% & 15.16\% \\
        COLA & 5.46\% & 10.00\% & 13.60\% \\
        MRPC & 5.21\% & 9.47\% & 15.63\% \\
        SST2 & 5.03\% & 9.41\% & 14.70\% \\
        QNLI & 5.37\% & 10.01\% & 14.07\% \\
        QQP & 5.80\% & 9.29\% & 14.97\% \\ \bottomrule 
    \end{tabular}
    \caption{The relative uncertainties ($\frac{\textrm{Std}(\textrm{MSE}_c)}{\textrm{Mean}(\textrm{MSE}_c)}$) using 3, 7, and 12 features to regress the 6 fine-tuning task performances.}
    \label{tab:std_to_mean_ratio_mse_c}
\end{table}

\subsection{Can the probing results distinguish the originating language models?}
\label{subsec:exp:distinguish-language-model}
Since the 25 models come from only 5 language models (RoBERTa, XLM, ALBERT, DeBERTa, XLNet), one may wonder if the ``model augmentation'' procedure ``shuffles'' the language models sufficiently -- if yes, then it would be hard to distinguish the originating language models. 

We use 5-class logistic regression from scikit-learn. For any combination of three features, we compute the accuracy following 5-fold cross validation. On all combinations of 3 features, the probing features can reach 0.0027 accuracy (sd=0.0109) better than the random features. This is statistically significant.\footnote{One-sample one-sided $t$-test on $\text{dof}=571,703$.} However, the maximum reachable accuracy is 0.08, whereas even a trivial predictor always outputting ``RoBERTa'' has an expected 0.24 accuracy (there are 6 RoBERTa models out of 25). The small accuracies show that our ``model augmentation'' procedure (\S \ref{subsec:pretrained-models}) produces sufficiently distinct models.

\section{Discussion}
\paragraph{Can probing results generalize to non-classification tasks?} All fine-tuning tasks and probing tasks in this paper are text-based classification problems. In the interpretable NLP literature, the probing analyses can also apply to other categories of deep neural networks including translation \citep{belinkov-etal-2017-neural,zhang-bowman-2018-language}. This generalization across different neural network is intuitive since probing examines the linguistic knowledge encoded in the representations. If a neural model encodes both rich syntactic information (as illustrated by high probing scores in Tense, SubjNum, etc.), and semantic information (as illustrated by high probing scores in BShift, SOMO, etc.) then we will not be surprised when observing that this neural model achieves a high BLEU score. That said, the extent to which probing results remain predictive for BLEU score needs further analysis, which we leave for future works.

\paragraph{Probing is computational-friendly} Compared to fine-tuning, probing evaluations require less computation. Fine-tuning the 6 GLUE tasks takes around 30 GPU hours in total, while probing the 7 tasks (all 12 layers) takes 0.7 GPU hours to cache and 1.3 CPU hours to probe. We elaborate the computational budgets in Appendix \ref{subsec:computation}. Probing is far more efficient because it does not need to change the parameters in the neural model, and we only need one pass through the neural model and cache the representations. Fine-tuning needs the gradients to update the parameters in the neural models. There are some methods to reduce the computation costs,\footnote{One approach involves using momentum to accelerate the convergence \citep{kingma2015adam,dozat2016incorporating} Alternatively, the memory usage can be reduced \citep{gomez2017reversible,pmlr-v97-behrmann19a}. Empirically, limiting the precisions can also accelerate optimization \citep{shin2021low}. Specially-designed structures including Adapters \citep{pmlr-v97-houlsby19a} and LoRA \citep{hu2022lora} are effective as well. Prefix tuning and prompt tuning are lightweight alternatives to fine-tuning \citep{he2022towards,le-scao-rush-2021-many,li-liang-2021-prefix}. \citet{liu_pre-train_2021} summarizes many approaches related to prompt.} and we note that probing is competitive as well, in terms of computational time.

\paragraph{Fine-tuning tasks need more specifications}
Currently, the most popular leaderboards for natural language understanding constitute fine-tuning tasks. A criticism towards these leaderboards is underspecification \citep{d2020underspecification} -- the short descriptions of the tasks can hardly be inclusive enough to specify the precise abilities required to complete the tasks. To further understand the underspecification problem, researchers recently developed probing datasets \citep{mccoy-etal-2019-right,warstadt-etal-2020-learning}. Probing results on these datasets have been (indirectly) used in developing deep neural models -- performance prediction is a more direct application.

\paragraph{Fine-grained evaluations improve transparency}
Leaderboard tasks should be customized to the users \citep{ethayarajh-jurafsky-2020-utility}. The diversity of probing datasets offers flexible choices to NLP researchers, supporting the diversified considerations to the consumers. Some recently proposed fine-grained leaderboards allow researchers to answer questions like ``where does model A outperform model B'' \citep{ma_dynaboard_2021,narayan_personalized_2021,ruder_xtreme-r_2021,liu-etal-2021-explainaboard}. The probing literature can provide many more datasets for building diverse leaderboards. 

\paragraph{Incorporating probing in model developments} 
While the developers are already busy, probing can still bring in benefits to the big model developments, mostly through a multi-dimensional feedback mechanism. The probing datasets introduce targeted knowledge that complement the training datasets of big NLP models. During developments, the model developers can select good model checkpoints to resume or proceed with the help of the probing evaluation scores.


\section{Conclusion}
This paper shows that analyzing probing results can be relevant to developing deep NLP models via predicting a proxy signal, i.e., fine-tuning performance. We show that as few as three probing accuracy scores can be useful to predict  fine-tuning results with RMSEs 40\% - 80\% smaller than baselines. This can dramatically improve the efficiency of deep learning pipelines. Given several ablation studies, we recommend MLP-20 and RandomForest-100 over other probing methods and show that the probing results from as few as 400 per class may still contain predictability. 
Probing analysis contain rich resources, and we show their results are closely related to fine-tuning performances. We call for further applications of probing into the developments of deep NLP models.

\section{Limitations}
The evaluation of large language models using only perplexity is uni-dimensional. Evaluations using fine-tuning tasks requires modifying many parameters, hence more costly than probing. Our paper aims at paving the path towards multidimensional evaluations of model parameters within computational budget, so instead of providing a fixed recipe (including fixing the probing datasets and specifying which layers to probe), we provide a general framework and use experiments to show the informativeness and potential utility of the probing results. While probing results are shown to be informative in our experiments, many other methods (e.g., LoRA and prefix-tuning) could optimize similar numbers of parameters. The empirical verifications of other methods are left to future work. Similarly, the problem settings considered in this paper are all classification problems. We believe this should generalize to other problem settings (e.g., BLEU score on sequential tasks), yet the empirical verifications are left to future work. 

\section*{Acknowledgement}
We thank the ARR reviewers for the comments. ZZ is supported by a Vector Institute Research Grant. FR is supported by a CIFAR AI Chair.

\bibliographystyle{acl_natbib}
\bibliography{bibliography}

\appendix

\section{Additional experimental details}
\label{subsec:computation}
\paragraph{On the effects of random seeds}
Literature has shown that the choice of random seeds could affect the fine-tuning results a lot \citep{dodge_fine-tuning_2020}, and we do not attempt to model this effect. Instead, by fixing the random seed \textit{before} any results are observed, we effectively control for the effect of random seeds in fine-tuning. An alternative experiment setup involves running multiple random seeds -- in this case, we will need a mixed-effect model instead of a simple regression model to factor out the effect of random seeds.

\paragraph{Computation budget for fine-tuning}
The computation budget for fine-tuning varies for the tasks (primarily due to the sizes of the datasets). The time usages for different language models do not differ much -- around 16 minutes for RTE, 17 minutes for MRPC, 45 minutes for COLA, 6.5 hours for SST2, 7 hours for QNLI, 18 hours for QQP. If we normalize by the sizes of the datasets, fine-tuning takes between 2 and 6 minutes of GPU time (on an RTX 6000 card; we refer to it as ``GPU'' henceforth) per 1,000 data samples.

\paragraph{Computation budget for probing}
Before probing, we cache the representations of SentEval data (taking around 60 hours GPU time) to avoid the feed-forward pass, which is the most time-consuming part. Note that we cached the whole SentEval data and then subsample 1,200 per class (around 1\%). Caching only the subsampled 1\% would take 40 minutes on GPU. 

The probing classifications take around 16 hours of CPU time (on an M1 chip; we refer to it as ``CPU'' henceforth) for each of the 25 models. This includes 7 (probing tasks) $\times$ 12 (layers) $\times$ 7 (probing methods) = 588 (probing configurations). 

Consider a configuration of one probing method, 12 layers, and 7 probing tasks. Acquiring 84 probing features would take 40 minutes (caching) + 80 minutes (probing), which is about two hours, where only 1/3 of the two hours need GPU. 

\paragraph{Computation in analysis}
For \S \ref{subsec:exp:best_3_features}, the feature selection procedure takes between 350 and 450 seconds for one fine-tuning task when running in RStudio on a laptop (with i7 CPU; we refer to it as RStudio henceforth). All 6 fine-tuning tasks take around 1 hour. Note that this procedure takes around 1/3 time on the M1 CPU.

For \S \ref{subsec:exp:ablation_probing_configuration}, the computation cost is 7 times that of \S \ref{subsec:exp:one_probing_task} to \S \ref{subsec:exp:best_3_features}, totaling around 7 hours.

For \S \ref{subsec:exp:ablation_smaller_probing_datasets}, the probing time is around 15 hours. Subsequent analysis time on RStudio equals \S \ref{subsec:exp:one_probing_task} to \S \ref{subsec:exp:best_3_features}, i.e., around 1 hour.

For \S \ref{subsec:exp:ablation-random-seed}, a Monte Carlo simulation for the uncertainty analysis takes around 15 minutes for all 6 fine-tuning tasks on CPU.

For \S \ref{subsec:exp:distinguish-language-model}, running through the features to find the top 3 for distinguishing the language models take 50 minutes on CPU.

\section{RMSE values}
Table \ref{tab:rmse-results-raw} shows the RMSE values complementing the RMSE reduction values in Table \ref{tab:rmse-results}. All values appear small in magnitude, but note that comparable scales of RMSE values can be achieved by the random features as well.

\begin{table}[t]
    \centering
    \resizebox{\linewidth}{!}{
    \begin{tabular}{r| r r r r r r r}
        \toprule 
         & RTE & COLA & MRPC & SST2 & QNLI  & QQP \\ \midrule 
        All layers one task (\S \ref{subsec:exp:one_probing_task}) \\
        BShift & .0353 & .0091 & .0160 & .0050 & .0040 & .0227 \\
        CoordInv & .0336 & .0054 & .0187 & .0044 & .0024 & .0218 \\
        ObjNum & .0427 & .0069 & .0174 & .0036 & .0034 & .0141 \\
        SOMO & .0274 & .0074 & .0173 & .0054 & .0040 & .0195 \\
        Tense & .0430 & .0092 & .0176 & .0055 & .0038 & .0100 \\
        SubjNum & .0489 & .0035 & .0176 & .0044 & .0026 & .0180 \\
        TreeDepth & .0378 & .0073 & .0207 & .0039 & .0031 & .0204 \\ \midrule 
        One layer per task (\S \ref{subsec:exp:one_layer_per_task}) & .0380 & .0068 & .0201 & .0048 & .0029 & .0383 \\ \midrule
        Only three features (\S \ref{subsec:exp:best_3_features}) & .0331 & .0053 & .0149 & .0028 & .0019 & .0125 \\
        \bottomrule
    \end{tabular}}
    \caption{RMSE values complementing the RMSE reduction values in Table \ref{tab:rmse-results}.}
    \label{tab:rmse-results-raw}
\end{table}

\end{document}